\documentclass{article}
\usepackage[preprint]{neurips_2026}

\usepackage[utf8]{inputenc}
\usepackage[T1]{fontenc}
\usepackage{hyperref}
\usepackage{url}
\usepackage{booktabs}
\usepackage{amsfonts}
\usepackage{nicefrac}
\usepackage{microtype}
\usepackage{xcolor}
\usepackage{graphicx}
\usepackage{amsmath}

\title{Position: Certifiable State Integrity Should Be Built from Local Validity, Not Global Scale}

\author{%
  Enzo Nicolás Spotorno \\
  Department of Informatics and Statistics\\
  Federal University of Santa Catarina\\
  Florianópolis, SC, Brazil \\
  \texttt{enzoniko@lisha.ufsc.br} \\
  \And
  Antônio Augusto Medeiros Fröhlich \\
  Department of Informatics and Statistics\\
  Federal University of Santa Catarina\\
  Florianópolis, SC, Brazil \\
  \texttt{guto@lisha.ufsc.br}
  \And
  João R. Campos \\
  Department of Engineering and Informatics\\
  University of Coimbra\\
  Coimbra, Portugal \\
  \texttt{jrcampos@dei.uc.pt}
}

\begin{document}
\maketitle

\begin{abstract}
Breakthroughs in language and vision have motivated increasingly general foundation models for time series and physical dynamics, where evidence is promising but less mature. In safety-critical Cyber-Physical Systems (CPS), globally parameterized dynamics models must remain valid through regime shifts, degradation, and lifecycle updates while supporting traceable assurance evidence. We argue that shared global parameters couple adaptation, frequency-sensitive fault representation, and verification evidence in ways that obstruct retained regime knowledge and bounded analysis. For systems whose operating conditions can be covered by a finite set of locally valid models, assurance evidence should be organized around those bounded models and their explicit composition, not around a single globally adapted model; increasing scale cannot supply explicit validity boundaries. HYDRA is a reference architecture for this position: frozen local specialists, runtime arbitration, and explicit coverage-gap monitoring. It illustrates how modular auditability and formal analysis of bounded components support Certifiable State Integrity, while whole-system certification, switching guarantees, and empirical validation remain separate obligations.
\end{abstract}

\section{Introduction}
\label{sec:intro}

Cyber-Physical Systems (CPS) increasingly operate across long lifecycles in which wear, maintenance, environmental changes, and operating-mode transitions alter the relationship between a physical asset and its digital representation \citep{gao2025survey,harkat2024cyber}. We use \textbf{State Integrity} for the continuous maintenance of a valid, evidence-backed correspondence between the physical system and the states represented by its computational model. The term is anchored in recent work on virtual twinning and state integrity under remote monitoring \citep{boche2025remote}, but is broadened here from communication-mediated estimation to lifecycle dynamics modeling. It entails three practical capabilities: reliable virtual sensing of unmeasured states \citep{ding2020secure}; drift-aware operation that distinguishes changes in operating conditions, degradation, and faults \citep{lu2018learning,poenaru2025improving}; and high-fidelity digital twinning that preserves decision-relevant dynamic features \citep{mohanraj2025digital}.

This usage is distinct from a Safety Integrity Level (SIL) or Automotive Safety Integrity Level (ASIL), which classify safety-related requirements and functions within particular standards, and from cybersecurity notions of data or message integrity. We call the narrower design objective studied in this paper \textbf{Certifiable State Integrity}: state integrity supported by evidence units whose assumptions, validity domains, interfaces, and failure responses can be inspected and incorporated into a safety case. The adjective \emph{certifiable} therefore denotes suitability for assurance analysis and bounded formal reasoning; it does not mean that a learning architecture, by itself, certifies a vehicle, controller, or safety function.

This distinction matters because functional-safety standards govern complete lifecycle processes and system-specific arguments, not isolated model forms. ISO~26262-9 addresses ASIL-oriented and safety-oriented analyses for road-vehicle electrical/electronic systems, including requirements decomposition, coexistence, dependent-failure analysis, and safety analyses \citep{iso26262part9}. IEC~61508-1 provides general lifecycle requirements for electrical/electronic/programmable electronic safety-related systems \citep{iec61508part1}. A modular model can help organize verification evidence, but frozen weights alone do not demonstrate Freedom from Interference, sufficient independence, ASIL decomposition, or compliance.

Foundation-model successes in language and vision have motivated general time-series models such as MOMENT, TimeGPT, and TimesFM \citep{liang2024foundation,goswami2024moment,garza2023timegpt,das2024decoder}, as well as neural operators for families of physical systems \citep{li2021fourier,lu2021learning}. Their transfer capabilities make them valuable candidates for CPS modeling. The open question is architectural: when the operating distribution changes throughout a safety-relevant lifecycle, should adaptation and evidence remain concentrated in one globally parameterized model?

Globally shared parameters create a plasticity--stability tension. Fine-tuning for a new regime can interfere with parameters that supported earlier regimes when gradients conflict \citep{mermillod2013stability,yu2020gradient}; recent time-series foundation-model experiments show that sequential fine-tuning can substantially degrade performance on earlier datasets \citep{karaouli2025time}. Forgetting is a possible outcome, not an inevitability, and depends on the update rule, data, capacity, and regularization. Test-time adaptation can reduce shift without maintaining a specialist library \citep{kim2025battling}, but it also changes the deployed artifact or its statistics and therefore requires an explicit assurance policy; Section~\ref{sec:altviews} treats it as a serious alternative. In a safety case, performance loss on a previously validated regime is consequential because it can invalidate earlier evidence, even though it is not itself a certification decision.

Three recurring challenges motivate our position. First, regimes separated by contact, topology, friction, or control-mode changes may be poorly represented by a single smooth global approximation. Second, some incipient faults and abrupt transitions have frequency-localized signatures that can be attenuated by objectives or architectures with spectral bias \citep{rahaman2019spectral,xu2025frequency}; this concern is signal- and architecture-dependent. Third, an end-to-end model may make evidence localization and revalidation expensive because a change can affect behavior across multiple regimes. Global parameter coupling is a common aggravating mechanism across these challenges, but it is neither their sole cause nor proof that every monolithic model will fail.

\textbf{Position.} For safety-critical CPS whose operating conditions can be covered by a finite set of locally valid models, assurance evidence should be organized around those bounded models and their explicit composition, not around a single globally adapted model; increasing model scale cannot supply explicit validity boundaries. Section~\ref{sec:theory} turns this scope condition into the operational Local Compressibility Condition. Gate behavior, interface contracts, dependent-failure analysis, and system-level validation remain separate obligations.

We use governance-inspired terms as compact labels, not as claims of organizational equivalence. A specialist's \emph{Sovereignty} is its frozen local parameterization; its \emph{Jurisdiction} is the validated domain $\Omega_k$; the \emph{Governor} is the blending or selection map; and the \emph{Constitution} is the set of shared physical and interface constraints. Section~\ref{sec:architecture} maps each label to a mathematical object.

HYDRA (Hierarchical uncertaintY-aware Dynamics with Runtime Arbitration) is a \emph{reference architecture} used to make the position concrete. It comprises a library of frozen local specialists, an uncertainty-aware Governor, and a coverage-gap monitor.


We make three contributions. \textbf{(C1)} We argue that, for the stated class of safety-critical CPS, the primary assurance unit should be a bounded local model with an explicit validity domain rather than a single globally adapted parameterization. \textbf{(C2)} We present HYDRA as a reference architecture that makes local models, composition interfaces, runtime arbitration, and coverage rejection explicit evidence objects, while separating those objects from system-level certification claims. \textbf{(C3)} We make the position technically refutable by stating the conditions required for local compressibility, convex composition, input-dependent Lipschitz analysis, hard-switching stability, and tail calibration, and by defining disconfirming tests for the central claims. Section~\ref{sec:theory} develops the challenges; Section~\ref{sec:architecture} presents the design; Section~\ref{sec:evidence} audits its assurance relevance; Section~\ref{sec:altviews} compares alternatives; and Section~\ref{sec:openproblems} gives the research agenda.

\section{Structural Challenges of Global Parameterization}
\label{sec:theory}

The following are architectural challenges, not impossibility theorems. They identify conditions under which local specialization may create more useful evidence boundaries than a single globally updated model.

\paragraph{Challenge 1: regime separation at transitions.}
A physical system can move between qualitatively different dynamics: high- to low-friction braking, free motion to rigid contact, or grid-connected to islanded operation. Global reduced representations can converge slowly for transport-dominated solution manifolds, whereas local reduced-order models and local bases can substantially improve approximation efficiency \citep{bachmayr2017kolmogorov,peherstorfer2014localized,amsallem2012local}. This motivates the following testable precondition.

\textbf{Local Compressibility Condition.} There must exist a finite collection of domains $\{\Omega_k\}_{k=1}^K$ and model classes $\{\mathcal{F}_k\}_{k=1}^K$ such that, under a declared capacity budget, each fitted specialist meets a held-out error tolerance on its domain and the collection covers the intended operational design domain. The condition does not exclude chaotic systems categorically; it fails whenever no practically sized partition and model family achieve the stated error and coverage targets. Appendix~\ref{app:theory} gives an empirical protocol based on held-out residual and capacity curves.

Local specialization is therefore proposed when local residuals saturate below the acceptance tolerance while the global model or an expanding local library does not. It is not justified merely because regimes have been named by an engineer.

\paragraph{Challenge 2: frequency-sensitive fault and transition evidence.}
Neural networks can exhibit spectral bias toward lower-frequency components \citep{rahaman2019spectral}. This does not imply that all faults are high-frequency or that every foundation model masks them. It does mean that validation should explicitly measure detection of frequency-localized signatures rather than infer it from aggregate prediction error. A deliberately narrow specialist may produce a detectable residual when operated outside its domain, but the residual is only \emph{evidence consistent with mismatch}; sensor faults, disturbances, and modeling error can produce the same observation. HYDRA therefore combines residual magnitude with gate uncertainty and persistence rather than calling a single spike an unambiguous regime label.

\paragraph{Challenge 3: reusable assurance evidence.}
When one parameter update can affect many operating regimes, previously collected evidence may require broad regression testing. A specialist library creates candidate evidence units: a local model, declared domain, dataset, error bound, and interface contract. This can localize impact analysis, but only if shared preprocessing, sensors, hardware, the Governor, and common training procedures are also analyzed. Gradient interference is a learning phenomenon; it is not the Freedom from Interference concept used in functional-safety analysis. Likewise, component modularity does not establish sufficient independence or justify ASIL decomposition.

Table~\ref{tab:challenge_examples} summarizes illustrative hypotheses. None of the local responses is guaranteed by architecture alone.

\begin{table}[t]
\centering
\small
\caption{Illustrative CPS transitions and the evidence sought from local specialization.}
\label{tab:challenge_examples}
\begin{tabular}{@{}p{1.7cm}p{2.6cm}p{3.2cm}p{3.4cm}@{}}
\toprule
Domain & Transition & Global-model concern & Local-specialization test \\ \midrule
Vehicle & dry to low friction & invalid intermediate estimate or delayed mismatch detection & residual detection delay and braking constraint violations \\
Robotics & free motion to contact & smoothing across a hybrid boundary & contact constraint violation and coverage-gap recall \\
Energy & strong grid to islanded & parameter averaging across topology change & frequency nadir and recovery under switching \\
Medical CPS & sedated to awake & delayed recognition under temporal averaging & alarm delay, false alarms, and subgroup calibration \\
\bottomrule
\end{tabular}
\end{table}

The proposed response is thus \emph{local validity isolation with explicit composition}. It addresses the shared-parameter aggravator, while leaving sensor ambiguity, dataset coverage, gate errors, common-cause failures, and closed-loop safety as separate obligations.

\section{Design Requirements and the HYDRA Reference Architecture}
\label{sec:architecture}

The preceding challenges motivate four design requirements for the class of systems targeted here. They are not universal necessities.

\paragraph{P1: local validity isolation.}
Each specialist $\mathcal{S}_k$ is frozen after validation and paired with a declared domain $\Omega_k$, training and validation provenance, and an error criterion. A claim is valid only within its stated domain and assumptions. Freezing prevents within-specialist online forgetting; it does not prevent integration changes elsewhere in the system.

\paragraph{P2: compositional analysis through explicit interfaces.}
Specialists expose common input/output contracts and local bounds, following the broader contract-based design principle that component guarantees are meaningful only under explicit assumptions and interfaces \citep{benveniste2018contracts}. Let $\mathcal{C}_k$ denote a validated local output or safety-relevant constraint set. If all active specialists produce outputs in a common convex set $\mathcal{C}$, then any convex blend also lies in $\mathcal{C}$. More generally, a blend lies in $\operatorname{Conv}(\cup_k\mathcal{C}_k)$, but that hull is not automatically safe. Robust positively invariant (RPI) zonotopes $\mathcal{Z}_k$ are one illustrative realization for linear or LPV closed-loop analyses.

\paragraph{P3: explicit coverage signaling.}
The system reports when current observations are poorly explained by the validated library. HYDRA uses a monitor $\mathcal{I}=(\mathcal{R},\mathcal{U},\mathcal{T})$: residual magnitude $\mathcal{R}$, uncertainty or dispersion in the Governor weights $\mathcal{U}$, and persistence or rate-of-change statistic $\mathcal{T}$. These signals require empirical calibration. Entropy is not automatically epistemic uncertainty, and a state-set exit indicates departure from a modeled envelope rather than a uniquely identified physical cause.

\paragraph{P4: controlled novelty-triggered growth.}
A new specialist is admitted only when held-out data show that the existing library cannot meet the declared tolerance and the candidate supplies non-redundant improvement. Sub-linear growth is a hypothesis to test against lifecycle exposure, not a consequence of greedy admission. Appendix~\ref{app:admission} states a reproducible protocol.

HYDRA is the tuple
\[
\mathcal{H}=(\mathcal{L},\mathcal{B},\mathcal{I}),\qquad
\mathcal{L}=\{(\mathcal{S}_k,\Omega_k,\mathcal{C}_k)\}_{k=1}^K,
\]
where $\mathcal{B}$ is the Governor and $\mathcal{I}$ is the coverage-gap monitor. Given an observation history $h_t$, the Governor produces $\boldsymbol{\pi}_t=\mathcal{B}(h_t)\in\Delta^{K-1}$ and
\begin{equation}
\hat y_t=\sum_{k=1}^K \pi_{k,t}\mathcal{S}_k(x_t).
\label{eq:blend}
\end{equation}
The weights can select a specialist, blend locally compatible specialists, or trigger fallback when the monitor rejects library coverage. Equation~\ref{eq:blend} is structurally analogous to a polytopic LPV representation, but nonlinear specialists and data-dependent gates do not inherit LPV guarantees automatically. Figure~\ref{fig:framework} illustrates, for an automotive case study, the corresponding data flow from the frozen specialist library through runtime arbitration and coverage-gap monitoring to the downstream controller.

\begin{table}[t]
\centering
\small
\caption{HYDRA nomenclature and the claims attached to each object.}
\label{tab:nomenclature}
\begin{tabular}{@{}p{2.4cm}p{3.4cm}p{2.1cm}p{2.8cm}@{}}
\toprule
Design role & Mathematical object & Shorthand & Evidence obligation \\ \midrule
local model & $\mathcal{S}_k$ on $\Omega_k$ & Sovereign specialist & held-out local validity \\
validity domain & $\Omega_k$ & Jurisdiction & coverage and boundary tests \\
shared constraints & $\Pi$ and interfaces & Constitution & contract and invariant checks \\
runtime composition & $\mathcal{B}$, $\boldsymbol{\pi}_t$ & Governor & gate calibration and timing \\
coverage rejection & $\mathcal{I}$ & Integrity Monitor & detection and false-alarm tests \\
library update & admission protocol & Accretion & novelty and regression evidence \\
\bottomrule
\end{tabular}
\end{table}

\subsection{Offline library construction}
\label{sec:offline}

Candidate regimes may be supplied by physics, operational knowledge, or data clustering. Each candidate is trained under shared physical/interface constraints $\Pi$ and evaluated on held-out local data. The admission test compares the candidate against the best existing admissible composition, records the marginal improvement, and reruns regression tests for the Governor and monitor. No existing specialist is retrained. This isolates model changes but does not eliminate shared-data or shared-toolchain dependencies.

\subsection{Online governance and monitoring}
\label{sec:integrity_triangle}

At runtime, residual likelihoods, discriminative gates, Bayesian model probabilities, or classical multiple-model filters can generate $\boldsymbol{\pi}_t$. A hard switch is appropriate for clearly separated hybrid modes; a soft blend is appropriate only when intermediate mixtures have a defensible physical interpretation and share a common safe output set. Conformal prediction may provide finite-sample marginal coverage under exchangeability or appropriate weighted/adaptive conditions \citep{barber2023conformal,zaffran2022adaptive}, but regime shift can violate those conditions. The coverage-gap monitor is therefore a separate rejection mechanism, not a consequence of conformal coverage.

A Simplex-style runtime assurance wrapper can route control to a separately developed fallback when the monitor rejects the high-performance channel \citep{bak2009system}. The wrapper can be used with either monolithic or modular AI. Its safety contribution depends on monitor coverage, switching latency, fallback capability, and independence/dependent-failure arguments; HYDRA does not assign SIL or ASIL levels to these elements. Figure~\ref{fig:simplex} illustrates this separation between the high-performance channel, the decision module, and the safety controller.

\section{Supporting Analyses and Assurance Connections}
\label{sec:evidence}

This section identifies what the architecture makes analyzable and what remains unproven.

\subsection{Compressibility and library growth}

If the Local Compressibility Condition holds, a local library may achieve a target error with lower per-model capacity than a global approximation \citep{peherstorfer2014localized,amsallem2012local}. The number of specialists remains an empirical quantity. Report $K(n)$ against the number $n$ of distinct lifecycle exposures, marginal error reduction per admission, memory and latency, and coverage. Evidence for sub-linear growth requires a fitted growth law with uncertainty and out-of-sample continuation; it cannot be inferred from an orthogonality heuristic alone.

\subsection{Compositional mathematical bounds}

For fixed or exogenous weights, local Lipschitz bounds compose:
\[
L_{\mathrm{blend}}\leq\sum_k\pi_kL_k\leq\max_k L_k.
\]
When the gate depends on the input, an additional gate-sensitivity term is required. Appendix~\ref{app:lipschitz} gives the derivation. Similarly, convex output containment is useful only after the relevant set has been shown to encode the downstream constraint. For hard switching, a standard average-dwell-time result can be stated using per-mode Lyapunov functions; it cannot be derived from specialist Lipschitz constants alone. Appendix~\ref{app:theory} states the special case and keeps neural soft-gating stability open.

\subsection{Assurance relevance without compliance claims}
\label{sec:assurance}

The architecture can expose evidence boundaries and mathematically specified verification targets in the spirit of Verified AI \citep{seshia2016verified}, but a standard-conformant safety case still depends on the item, hazards, requirements, development process, hardware, software, and organizational evidence. Table~\ref{tab:assurance} shows an evidence-gap map.

\begin{table}[t]
\centering
\small
\caption{Assurance relevance of local specialization and evidence that remains necessary.}
\label{tab:assurance}
\begin{tabular}{@{}p{2.5cm}p{3.6cm}p{4.2cm}@{}}
\toprule
Assurance concern & Potential architectural contribution & Evidence not supplied by HYDRA \\ \midrule
component boundaries & local model/domain/version records & safety requirements allocation and bidirectional traceability \\
Freedom from Interference & candidate spatial and change boundaries & timing, memory, communication, and resource interference analysis \\
dependent failures & explicit shared Governor and interfaces & common-cause analysis for sensors, preprocessing, hardware, data, and tools \\
requirements decomposition & separable evidence packages may aid allocation & sufficient independence and valid decomposition argument \\
formal bounds & local set and sensitivity analyses & closed-loop assumptions, gate behavior, numerical validity, and system integration \\
runtime assurance & explicit rejection and fallback interface & monitor coverage, switching latency, fallback authority, and verification \\
\bottomrule
\end{tabular}
\end{table}

Logging the active specialist improves observability but is not requirements traceability. A Lipschitz bound can support a particular robustness argument but is not, by itself, an ASIL requirement. Frozen specialists reduce one source of change propagation but do not establish Freedom from Interference or sufficient independence \citep{iso26262part9,iec61508part1}.

\subsection{Tail calibration of monitor thresholds}
\label{sec:evt}

Extreme Value Theory (EVT) offers a parametric approximation for high residual tails. For a sufficiently high threshold $u$, exceedances may be approximated by a Generalized Pareto Distribution (GPD):
\begin{equation}
P(\mathcal{R}>\tau\mid\mathcal{R}>u)
\approx
\left(1+\xi\frac{\tau-u}{\sigma}\right)^{-1/\xi}.
\label{eq:gpd}
\end{equation}
The resulting threshold is a fitted operating point, not a certified false-alarm guarantee. Validation must account for threshold selection, dependence, model fit, parameter uncertainty, and shift between calibration and deployment. Appendix~\ref{app:evt} gives a conservative protocol using held-out evaluation and uncertainty intervals.

\subsection{Falsifiable evaluation plan}

The evaluation vocabulary follows the adjacent position of \citet{dai2026category} in using Time-to-Recovery and Intervention Recall for adaptation after structural change. We retain those attributions, add Time-to-Uncertainty for coverage rejection, and place all three inside a safety-oriented evaluation that also measures false interventions, fallback occupancy, and constraint violations.

\begin{table}[t]
\centering
\small
\caption{Claims, disconfirming evidence, and minimum reported measurements.}
\label{tab:falsifiability}
\begin{tabular}{@{}p{2.4cm}p{3.7cm}p{4.1cm}@{}}
\toprule
Claim & Disconfirming result & Required measurements \\ \midrule
local compressibility & local capacity or library size fails to improve the error--cost frontier & held-out residual vs.\ capacity, $K$, memory, latency \\
EVT calibration & observed tail error lies outside uncertainty targets or is unstable across thresholds & exceedance diagnostics, confidence intervals, held-out false-alarm rate \\
coverage-gap detection & novel regimes are accepted or known regimes are repeatedly rejected & intervention recall, false-alarm rate, time-to-uncertainty \\
availability and recovery & fallback dominates operation or recovery is too slow for the control envelope & time-to-recovery, fallback occupancy, constraint violations \\
\bottomrule
\end{tabular}
\end{table}

\section{Alternative Perspectives and Boundary Conditions}
\label{sec:altviews}

Local specialization is one point in a broader design space. Table~\ref{tab:alternatives} compares four alternatives.

\begin{table}[t]
\centering
\small
\caption{Alternative approaches and their main assurance question.}
\label{tab:alternatives}
\begin{tabular}{@{}p{2.2cm}p{3.0cm}p{3.0cm}p{2.7cm}@{}}
\toprule
Approach & Native strength & Boundary relative to HYDRA & Assurance question \\ \midrule
test-time adaptation & adapts one deployed model to shift & may avoid library growth but mutates model/statistics & what evidence remains valid after each update? \\
mixture of experts & scalable conditional computation & experts need not map to validated physical regimes & are routing and expert domains semantically testable? \\
filtered monolith + Simplex & simple high-performance channel with fallback & can isolate runtime risk without local models & can monitor and fallback cover unsafe outputs in time? \\
IMM/MMAE & established multiple-model Bayesian filtering & often assumes specified mode models and transition laws & are mode set, probabilities, and residual models adequate? \\
\bottomrule
\end{tabular}
\end{table}

\paragraph{Test-time adaptation.}
TTA can update features, normalization statistics, prompts, or parameters using incoming data; time-series methods such as TAFAS demonstrate that partially observed ground truth can support adaptation under continuous distribution shift \citep{kim2025battling}. It is attractive when shifts are gradual and storage is constrained. It challenges assurance because the deployed artifact or its internal statistics change, adaptation errors can accumulate, and rollback criteria must be defined. A viable comparison should hold compute and data constant and measure both current-regime gain and regression on previously validated regimes.

\paragraph{Mixture of experts.}
Sparse MoE architectures provide conditional capacity and can reduce inference cost \citep{shazeer2017,yang2025mixture}. They are compatible with our position when expert domains, routing behavior, and update policies become explicit assurance objects. They differ when experts emerge only as latent computational shards: specialization alone does not yield physical jurisdiction, calibrated rejection, or reusable local evidence.

\paragraph{Filtered monolith and Simplex.}
A fixed global model inside a runtime-assurance wrapper may be preferable when a capable safety controller and reliable decision module already exist. Simplex separates the high-performance and high-assurance channels regardless of whether the former is monolithic or modular \citep{bak2009system}. HYDRA's added value must therefore be demonstrated through earlier coverage-gap detection, easier revalidation, or improved availability, rather than assumed from the wrapper.

\paragraph{IMM and MMAE.}
Interacting Multiple Model and Multiple-Model Adaptive Estimation methods already combine mode-conditioned filters using residual-based probabilities \citep{blom1988imm,barshalom2001estimation,imm2022,mmae2011}. HYDRA should be understood as extending this design pattern to learned local operators, explicit coverage rejection, and assurance-oriented evidence packages. IMM/MMAE are strong baselines and may remain preferable when reliable parametric mode models exist.

\section{Open Problems and Research Directions}
\label{sec:openproblems}

The position becomes actionable when its gaps can be turned into tests.

\paragraph{1. Discovering and limiting jurisdictions.}
\emph{Gap:} a greedy residual criterion may produce redundant or order-dependent partitions. \emph{Test:} compare physics-defined, clustering-based, and greedy partitions under identical capacity; report held-out error, coverage, $K$, admission stability, and computational cost. Reject P4 when library growth or marginal utility does not meet a preregistered target.

\paragraph{2. Stability under switching and blending.}
\emph{Gap:} local model validity does not imply stable switching or stable closed-loop control. \emph{Test:} for hard switching, identify per-mode Lyapunov functions and verify a cross-mode factor $\mu$ and average dwell time $\tau_a>\ln(\mu)/\alpha$; then test perturbations and monitor delays. For soft neural gates, estimate gate sensitivity and perform closed-loop reachability or falsification. Appendix~\ref{app:theory} states only the hard-switching special case.

\paragraph{3. Calibrated coverage-gap detection.}
\emph{Gap:} residuals and gate entropy can be miscalibrated under dependence and drift. \emph{Test:} introduce held-out known regimes, boundary mixtures, sensor faults, disturbances, and truly novel regimes; report intervention recall, time-to-uncertainty, false alarms, and calibration by subgroup and regime duration.

\paragraph{4. Availability-preserving fallback and recovery.}
\emph{Gap:} conservative rejection can make the system safe but unavailable. \emph{Test:} jointly measure constraint violations, fallback occupancy, time-to-recovery, control performance, and unsafe re-entry across novelty severity. Compare against TTA, IMM/MMAE, and a filtered monolith.

\paragraph{5. Assurance integration and independence.}
\emph{Gap:} modular learning components can still share sensors, hardware, preprocessing, data, tools, and the Governor. \emph{Test:} build a system-specific assurance case that traces requirements to local and shared evidence; perform interference and dependent-failure analyses; inject common-cause faults; and measure the revalidation impact of adding or replacing one specialist. The hypothesis is reduced evidence-change scope, not automatic compliance.

Candidate implementations such as neural operators, kernel models, KANs, or parameter-space blending should be compared only after these architecture-level obligations are fixed; no particular model family is essential to the position.

\section{Conclusion}
\label{sec:conclusion}

Globally parameterized models remain valuable baselines and may be the right choice when operating distributions are stable, updates are tightly controlled, or a separate assurance wrapper provides adequate coverage. Our claim is that lifecycle CPS with compressible local regimes may benefit from compositional local specialization: frozen specialists can preserve local evidence, an explicit Governor can make composition inspectable, and a coverage-gap monitor can turn unsupported operation into a testable event.

HYDRA is a reference architecture for studying that claim. Its value will be established only if it outperforms strong alternatives on retained validity, coverage-gap detection, closed-loop safety, availability, library growth, and evidence-change scope. The immediate research direction is therefore to test whether explicit local validity and composition produce better assurance evidence under realistic lifecycle change.

\bibliographystyle{plainnat}
\bibliography{CURRENT_ref}

\appendix

\section{Greedy Library Admission Protocol}
\label{app:admission}

For a candidate dataset $D_c$, first evaluate every existing specialist and every permitted composition on held-out data. Let $E_{\mathcal{L}}(D_c)$ be the best admissible validation error and $E_c(D_c)$ the candidate's error at the same capacity-accounting rule. Admit the candidate only if: (i) $E_{\mathcal{L}}(D_c)$ exceeds the declared tolerance; (ii) $E_c(D_c)$ meets the tolerance; (iii) the improvement exceeds a preregistered margin with uncertainty; and (iv) the candidate passes interface, shared-constraint, and regression tests. Record the candidate's domain description, data provenance, failure cases, and computational cost. Re-evaluate the Governor and monitor after admission. This protocol is inspired by greedy reduced-basis and orthogonal matching-pursuit ideas \citep{pati1993orthogonal, ji2025s2gptpinn, chen2024gptpinn}; it does not guarantee a minimal or sub-linearly growing library.

\section{Governor Inference Methods}
\label{app:governor}

Possible Governors include Bayesian model probabilities, recursive filters, discriminative gates, and residual-based deterministic scores. Bayesian or Dirichlet formulations can represent uncertainty over $\boldsymbol{\pi}$; IMM-style filters can encode mode-transition dynamics; deterministic exponential smoothing can meet tight latency budgets. These methods are not interchangeable: their calibration, temporal assumptions, and worst-case execution time must be validated for the application. Sparse weights can support hard regime selection, whereas dense weights should be used only where mixtures have a defensible interpretation. Claims of control-rate feasibility require measured worst-case latency on target hardware, including all active specialists and monitor logic.

\section{Runtime Assurance and Simplex Integration}
\label{app:simplex}

A Simplex-style arrangement contains a high-performance channel, a verified or otherwise justified safety controller, and a decision module \citep{bak2009system}. HYDRA may occupy the high-performance channel. When the coverage monitor rejects the library, the decision module can transfer authority to the safety controller. This architectural separation does not assign SIL or ASIL levels, prove Freedom from Interference, or establish sufficient independence. The safety case must analyze shared sensors, power, processors, communication, timing, memory, and failure modes; verify that the decision module detects hazards early enough; and show that the fallback can maintain the relevant safe set.

\IfFileExists{simplex_architecture1.png}{
\begin{figure}[h]
\centering
\includegraphics[width=0.58\linewidth]{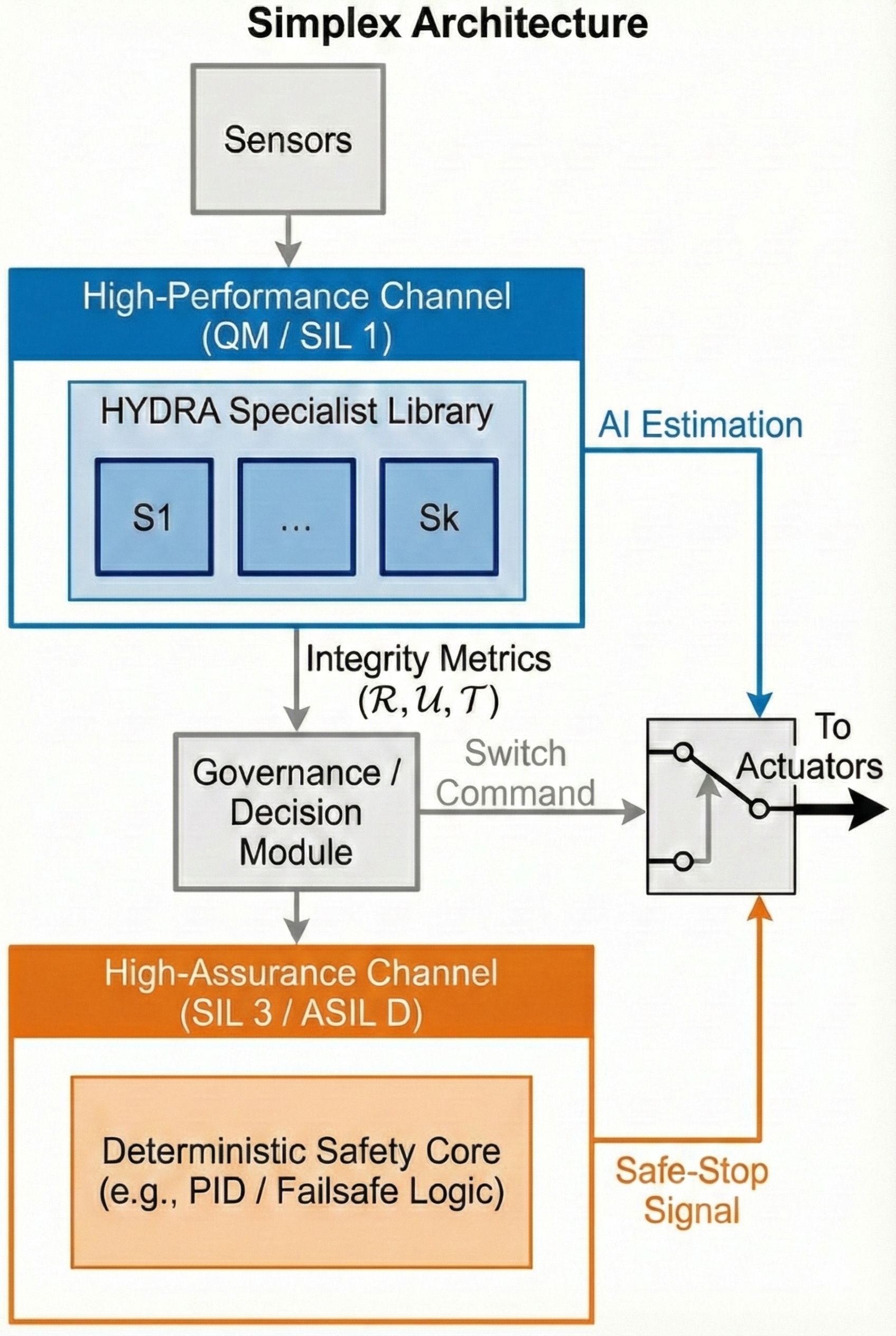}
\caption{Illustrative Simplex integration. Assurance claims depend on the implemented decision module, fallback, interfaces, and independence analysis.}
\label{fig:simplex}
\end{figure}
}{}

\section{Illustrative Vehicle Scenario}
\label{app:vehiclecasedetails}

Consider a vehicle moving from dry asphalt to a low-friction patch \citep{karyotakis2025braking}. A dry-road specialist may develop a large wheel-dynamics residual; the Governor may shift weight to a validated low-friction specialist; and, during unresolved ambiguity, the monitor may request fallback. This is a hypothesis, not a guaranteed sequence. A valid experiment must include sensor noise, split-friction conditions, previously unseen tires, gradual friction changes, and faults that mimic regime mismatch. It should compare HYDRA with a fixed global model, TTA, IMM/MMAE, and each method inside the same runtime-assurance wrapper. Primary outcomes are constraint violations, braking performance, time-to-uncertainty, false interventions, and recovery.

\IfFileExists{framework.png}{
\begin{figure}[h]
\centering
\includegraphics[width=\linewidth]{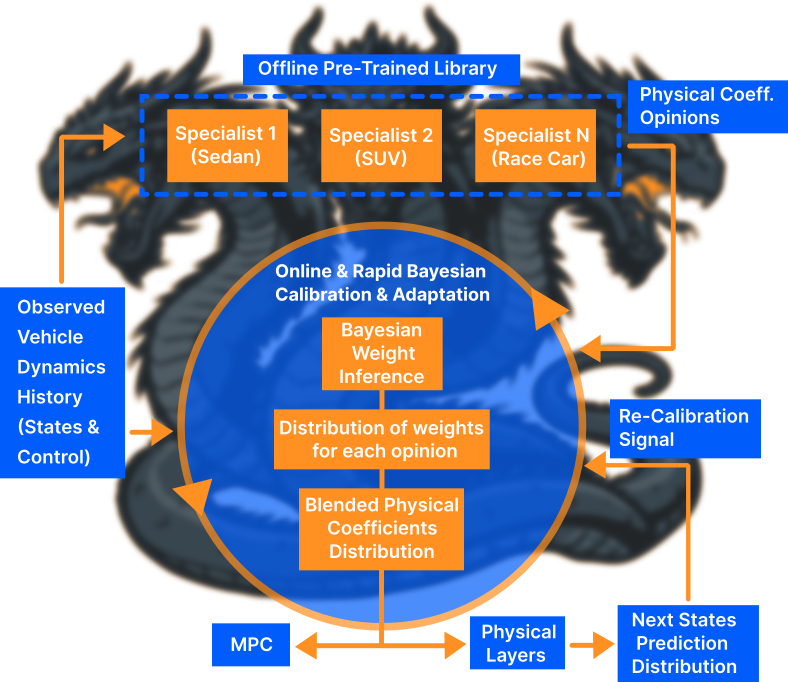}
\caption{Illustrative HYDRA data flow from frozen specialists through the Governor and monitor to a downstream controller.}
\label{fig:framework}
\end{figure}
}{}

\section{Local Constraint Sets and an RPI Example}
\label{sec:sets}

Let $\mathcal{C}_k$ denote the validated output or safety-relevant constraint set for specialist $k$. If the active specialists all satisfy $\mathcal{S}_k(x)\in\mathcal{C}$ for the same convex set $\mathcal{C}$, Equation~\ref{eq:blend} gives $\hat y(x)\in\mathcal{C}$. If the sets differ, convex blending yields only
\[
\hat y(x)\in\operatorname{Conv}\left(\bigcup_{k:\pi_k>0}\mathcal{C}_k\right),
\]
and a separate argument must show that this hull is acceptable.

For a linear or LPV closed-loop specialist, $\mathcal{C}_k$ may be instantiated by an RPI zonotope $\mathcal{Z}_k$ computed under bounded disturbances \citep{morato2023predictive,mayne2000constrained}. A measured or estimated state outside the relevant $\mathcal{Z}_k$ is evidence that the local envelope assumptions no longer hold. It is not an unambiguous diagnosis: estimator error, disturbance-bound violation, or sensor failure are alternative causes. Nonlinear learned specialists require domain-appropriate reachability or conservative enclosure methods.

\section{Lipschitz Composition Bounds}
\label{app:lipschitz}

Assume a common domain $D$ and $\|\mathcal{S}_k(x)-\mathcal{S}_k(x')\|\leq L_k\|x-x'\|$ on $D$.

\paragraph{Fixed or exogenous weights.}
For $\boldsymbol{\pi}\in\Delta^{K-1}$ independent of $x$,
\begin{align}
\|\hat y(x)-\hat y(x')\|
&\leq \sum_k\pi_k\|\mathcal{S}_k(x)-\mathcal{S}_k(x')\|\\
&\leq \left(\sum_k\pi_kL_k\right)\|x-x'\|
\leq \max_k L_k\|x-x'\|.
\label{eq:lipschitz}
\end{align}
Equality with $\max_k L_k$ is attainable by placing all weight on a specialist with the largest constant.

\paragraph{Input-dependent gate.}
Let $\hat y(x)=\sum_k\pi_k(x)\mathcal{S}_k(x)$. If $\|\boldsymbol{\pi}(x)-\boldsymbol{\pi}(x')\|_1\leq L_\pi\|x-x'\|$, $\|\mathcal{S}_k(x)\|\leq B$ on $D$, and $L_{\max}=\max_kL_k$, then
\begin{align}
\|\hat y(x)-\hat y(x')\|
&\leq \sum_k\pi_k(x)\|\mathcal{S}_k(x)-\mathcal{S}_k(x')\|\\
&\quad+\sum_k|\pi_k(x)-\pi_k(x')|\,\|\mathcal{S}_k(x')\|\\
&\leq (L_{\max}+BL_\pi)\|x-x'\|.
\end{align}
Thus the fixed-weight bound cannot be reused for an input-dependent Governor without a gate term. Domain boundaries, discontinuous hard switches, and coordinate transformations require separate treatment.

\section{EVT Threshold Calibration}
\label{app:evt}

Given normal-operation residuals, select candidate high thresholds using threshold-stability and mean-residual-life diagnostics, fit a GPD to exceedances, and propagate parameter uncertainty to the target quantile. Serial dependence should be handled by declustering or a dependence-aware method. Evaluate the chosen threshold once on untouched held-out sequences and report the empirical false-alarm rate with a binomial or dependence-appropriate uncertainty interval. Goodness-of-fit should combine graphical diagnostics, threshold sensitivity, and tests whose calibration accounts for fitted parameters; a naive Kolmogorov--Smirnov test against the fitted sample is insufficient. Repeat across regimes and realistic shifts. If tail fit is unstable or the deployment distribution changes, use the monitor threshold only as an empirical operating point and trigger recalibration or a conservative fallback policy.

\section{Theoretical Grounding}
\label{app:theory}

\subsection*{H.1 Polytopic LPV analogy}

A polytopic LPV model represents dynamics as a convex combination of vertex systems indexed by a scheduling variable \citep{balas2002linear,morato2023predictive}. Equation~\ref{eq:blend} has a similar algebraic form. This analogy suggests candidate tools, but HYDRA does not automatically inherit LPV robustness results because its specialists may be nonlinear, its gate may depend on uncertain observations, and its local coordinates may differ. Any use of LPV analysis must verify common coordinates, admissible scheduling rates, local-domain coverage, and closed-loop assumptions.

\subsection*{H.2 Hard-switching stability special case}

Consider hard switching among modes with Lyapunov functions $V_k(x)$ satisfying, while mode $k$ is active,
\[
V_k(x(t))\leq e^{-\alpha(t-t_0)}V_k(x(t_0)),\qquad \alpha>0,
\]
and at switching instants a uniform comparison $V_j(x)\leq\mu V_i(x)$ for all admissible transitions, with $\mu\geq1$. Standard switched-systems reasoning implies exponential stability under an average dwell time satisfying
\[
\tau_a>\frac{\ln\mu}{\alpha},
\]
up to the usual chatter bound \citep{liberzon1999basic}. The quantities $\alpha$ and $\mu$ must come from mode and transition analysis; they do not follow from specialist Lipschitz constants. This proposition covers neither arbitrary soft blending nor learned neural closed loops.

\subsection*{H.3 Local Compressibility Condition}

For a solution set $\mathcal{M}$ in a normed space, the Kolmogorov $n$-width
\[
d_n(\mathcal{M})=\inf_{\dim(V)=n}\sup_{x\in\mathcal{M}}\inf_{v\in V}\|x-v\|
\]
measures the best worst-case linear approximation error. Transport-dominated manifolds can have slow global decay, motivating localized bases and local reduced-order models \citep{bachmayr2017kolmogorov,peherstorfer2014localized,amsallem2012local}. HYDRA does not require exponential decay as a theorem. Its practical condition is assessed by plotting held-out error against model capacity for each proposed domain, together with total library memory and latency. A partition is supported only when local curves achieve the preregistered tolerance within budget and unseen in-domain data retain coverage. Persistent residual saturation, unstable partitions, or unbounded library growth falsify the condition for the chosen representation and operating domain.

\end{document}